\documentclass[11pt, a4paper]{article}
\usepackage{tacl2021v1}

\usepackage{times,latexsym}
\usepackage{tipa}
\usepackage{hyperref}
\usepackage[T1]{fontenc}
\usepackage{amssymb}
\usepackage{CJKutf8}
\usepackage{microtype}
\usepackage{booktabs}
\usepackage{multirow}
\usepackage{amsmath}
\usepackage{multirow, hhline, graphicx, helvet}
\mathchardef\mhyphen="2D

\usepackage{makecell}
\usepackage{xspace, mfirstuc, tabulary}
\newcommand{\base}{\textsc{base}\xspace}
\newcommand{\labm}{\textsc{label}$_m$\xspace}
\newcommand{\labs}{\textsc{label}$_s$\xspace}
\newcommand{\labsr}{\textsc{label}$_{sr}$\xspace}
\newcommand{\labmr}{\textsc{label}$_{mr}$\xspace}

\newcommand{\dall}{\textsc{all}\xspace}
\newcommand{\dhigh}{\textsc{high}\xspace}
\newcommand{\dmid}{\textsc{mid}\xspace}
\newcommand{\dallf}{\textsc{all+freq}\xspace}

\newcommand{\MODA}{\textsc{model\scriptsize{[-pinyin]}}\xspace}
\newcommand{\MODB}{\textsc{model\scriptsize{[+pinyin]}}\xspace}

\newcommand{\regular}{\emph{regular}\xspace}
\newcommand{\alliterating}{\emph{alliterating}\xspace}
\newcommand{\rhyming}{\emph{rhyming}\xspace}
\newcommand{\rad}{\emph{irregular}\xspace}
\newcommand{\sem}{\emph{semantic}\xspace}

\newcommand{\std}[1]{{\scriptsize $\pm$#1}}

\usepackage{todonotes, color}
\newcommand{\Note}[2]{} 
\newcommand{\SideNote}[2]{}
\renewcommand{\Note}[2]{\todo[color=#1,size=\small, inline=true]{#2}} 
\renewcommand{\SideNote}[2]{\todo[color=#1,size=\small]{#2}} %

\title{Evaluating Transformer Models and Human Behaviors on Chinese Character Naming}

\author{Xiaomeng Ma \\
  The Graduate Center, CUNY \\
  New York, USA\\
  \texttt{xma3@gradcenter.cuny.edu} \\\And
  Lingyu Gao \\
  Toyota Technological Institute at Chicago \\
Chicago, USA\\
  \texttt{lygao@ttic.edu} \\}
\begin{document}
\maketitle
\begin{abstract}
Neural network models have been proposed to explain the grapheme-phoneme mapping process in humans for many alphabet languages. These models not only successfully learned the correspondence of the letter strings and their pronunciation, but also captured human behavior in nonce word naming tasks. How would the neural models perform for a non-alphabet language (e.g., Chinese) unknown character task? How well would the model capture human behavior?  In this study, we first collect human speakers' answers on unknown character naming tasks and then evaluate a set of transformer models by comparing their performances with human behaviors on an unknown Chinese character naming task. We found that the models and humans behaved very similarly, that they had similar accuracy distribution for each character, and had a substantial overlap in answers. In addition, the models' answers are highly correlated with humans' answers. These results suggested that the transformer models can well capture human's character naming behavior. \footnote{The code and data for this paper can be found at: \href{https://github.com/xiaomeng-ma/Chinese-Character-Naming}{https://github.com/xiaomeng-ma/Chinese-Character-Naming}.}
\end{abstract}
\begin{CJK*}{UTF8}{bsmi}
\section{Introduction}
\label{section:!}

Many aspects of language can be characterized as quasi-regular: the relationship between inputs and outputs is systematic but allow many exceptions. The grapheme-phoneme mapping is an example of such quasi-regularity. For example, the letter string `\textit{-ave}' in English is regularly pronounced as /e\textsci v/ in \textsc{gave}, \textsc{save}, with the exception of /\ae v/ in \textsc{have}. And human speakers can easily grasp both patterns, e.g., in a nonce word naming experiment, most speakers pronounced the word \textsc{tave} as /te\textsci v/, while some pronounced it as /t\ae v/ \cite{glushko1979organization}. 

To explain the grapheme-phoneme mapping process, many models have been proposed, among which the Dual Route Cascaded (DRC) model and the connectionist model are the two most influential yet opposite models. The DRC model \citep{coltheart2001drc, coltheart1978lexical} proposes that the grapheme-phoneme mapping is implemented in two separate routes: a lexical route that directly maps the word's spelling to its pronunciation through a dictionary-like lookup procedure\footnote{The lexical route is usually applied to sight words (e.g., `of', `and') and words that don't follow grapheme-phoneme correspondence rules (e.g., `colonel').}, and a non-lexical route that applies the grapheme-phoneme corresponding `rules' to convert the letters to their corresponding pronunciation. The implementation of the DRC model requires domain-specific knowledge, such as spelling to sound rules. In contrast, the connectionist model \citep{seidenberg1989distributed, plaut1996understanding} proposed that a word's pronunciation is generated through a neural network that takes the orthographic representation as the input and outputs the phonological representation, which does not require specific knowledge of grapheme-phoneme correspondence rules. Both models can explain various behaviors in word identification, such as the faster identification of frequent words compared to infrequent ones. Therefore, there is still an ongoing debate about which model better captures the grapheme-phoneme mapping process. 

However, most of these two models were tested on alphabetic languages (e.g., English and German), and it is still unclear how would these models be generalized to a non-alphabetic language, such as Chinese. The DRC model seems to be unfit for Chinese because there are no regularities in Chinese that can be defined as grapheme-phoneme corresponding rules \citep{yang2009simulating}. In addition, \citet{coltheart2001drc} asserted that ``the Chinese, Japanese and Korean writing systems are structurally so different from the English writing system, that a model like the DRC model would simply not be applicable.'' (p.236). Thus the connectionist model is the only candidate. The majority (81\%) of Chinese characters are phono-semantic compounds \cite{li1993analysis}, which consist of a phonetic radical that contains pronunciation information (denoted by pinyin),\footnote{Chinese characters use pinyin to represent the pronunciation. The pinyin system consists of 24 syllable initials (mostly contain a consonant), 34 syllable finals (mostly contain a vowel or vowels), and 4 tones.} and a semantic radical that contains semantic information.\footnote{The phonetic radical and semantic radical are mutually exclusive, and they are defined in the ancient Chinese dictionary 《說文解字》\textit{`Shuowen Jiezi'}.} For example, for the character 晴 ($<$qing2$>$ `sunny'), the left side 日 ($<$ri4$>$, `sun') is the semantic radical, and the right side 青 ($<$qing1$>$, `blue') is the phonetic radical. While the phonetic radical does not contain componential information about the pronunciation, e.g., the first part of the phonetic radical does not represent the first phoneme (e.g., consonant)/syllable onset as letter strings, the relationship between the phonetic radical's pinyin and the character's pinyin is also quasi-regular. Ignoring the tonal differences, the character's pinyin can be categorized into 4 types \cite{fang1986consistency}: \regular, the same as the phonetic radical's pinyin; \alliterating, deviating in the syllable final; \rhyming, deviating in the syllable onset; and \rad, varying in both syllable onset and final (see Table \ref{tab:0} for examples). The process to pronounce an unknown character involves two steps, where the first step is to identify the phonetic radical, and the second step is to apply the regularity pattern of the pinyin. However, there are no reliable cues to identify the phonetic radical, and the regularity patterns are quite arbitrary \citep{yang2009simulating}. How do Chinese speakers name an unknown character, and how well can the neural models capture the Chinese speakers' behaviors?

\begin{table}[t!]
\centering
\small
\begin{tabular}{p{0.2\linewidth}|p{0.65\linewidth}}
\toprule
 & Example characters \\
\midrule
\regular & 清, 情, 圊, 晴 -- $<$\textbf{qing}$>$ \\
\alliterating & 倩，輤 $<$\textbf{q}ian$>$\\
\rhyming & 精, 靖, 菁 -- $<$j\textbf{ing}$>$ \\
\rad & 猜 $<$cai$>$, 靚 $<$liang$>$, 靛 $<$dian$>$ \\ \bottomrule
\end{tabular}
\caption{\small Examples of characters with the phonetic radical 青 $<$qing$>$, sorted into different regularity types. Syllable onsets and finals are \textbf{bold} when they are the same with the phonetic radical.}
\label{tab:0}
\end{table}

In our study, we first collected human speakers' answers on unknown character naming, since there is no study investigating how Chinese adults read unknown characters.\footnote{Previous studies have focused on children's behavior on unknown character naming and found that children made errors in identifying the incorrect phonetic radical, as well as applying the incorrect regularity pattern \cite{lam2008exploratory, lam2014elaborating}.} We then trained a set of sequence-to-sequence transformer models with different settings on 4,281 phono-semantic characters. Neither human speakers nor models can name the unknown characters accurately, but the transformers have a slightly better average accuracy (47.4\%) than the human speakers (45.3\%). We then evaluated how closely the results of our aggregated transformers matched those of the human participants, in aspects of the variety of answer types and answer overlaps. In general, both the transformers and human speakers are able to identify the phonetic radical correctly and apply all 4 types of regularities to infer the pinyin, and the transformer models show a high correlation with human data in the proportion of each regularity type. In addition, 
there is a considerable amount of agreement between the answers generated by our models and those given by humans.
Our results demonstrate that transformer models can well capture human behavior in unknown Chinese character naming. 

\section{Related Work and Current Study}
Skilled Chinese readers make use of the phonetic radicals to name characters \cite{chen1996functional, zhou1999sublexical, ding2004nature}, and previous studies measured how phonetic radicals influence character naming in two ways: regularity and consistency \cite{fang1986consistency, hue1992recognition, hsu2009orthographic}. The regularity is exemplified in Table~\ref{tab:0}, and the consistency is defined as the number of characters that share the same phonetic radicals and pinyin. For example, there are 12 characters sharing the phonetic radical 青 $<$qing$>$ in Table~\ref{tab:0}, among which 3 characters (精, 靖, 靖) have the same pinyin $<$jing$>$, so the consistency score for these characters is 0.25 (3/12). Many studies have found \textit{regularity and consistency effects} for human speakers %
- the \regular and more consistent characters are named faster and more accurately, 
and these effects are stronger for low-frequency characters than high-frequency ones \cite{lien1985consistency, liu2003regularity, tsai2005consistency}. 

Previous studies of Chinese character modeling with phonetic radicals as inputs have successfully simulated the regularity effect and consistency effect. \citet{yang2009simulating} trained a feed-forward network on 4,468 Chinese characters and tested the model on 120 characters (seen in the training). The input to the model includes the character's radicals and radicals' positions (e.g., left-right, up-down).\footnote{There are 10 different Chinese character structures clustered by the arrangement of the character radicals, e.g., left-right (日+青=晴), top-down (相+心=想), and enclosure (口+或=國). The left-right structure is the most common type (71\%) \cite{hsiao2006analysis}.} The output of the model is the phonological features (e.g., stop, lateral) of the character's pinyin. They also measured the human speakers' response latency\footnote{Response latency measures the response speed, usually in milliseconds.} on each of the 120 test characters. By comparing the human speakers' response latency and the model's sum squared error, they found very similar regularity and consistency effects. In addition, \citet{hsiao2004connectionist,hsiao2005differences} trained a feed-forward model on 2,159 left-right structured characters, with each character appearing according to its log token frequency. The input included each character's radicals, and the output was the character's pinyin. They analyzed the training accuracy of the model and found the model's sum squared errors lower for the \regular characters, which successfully simulated the regularity effect. 

The regularity and consistency effect revealed that both human speakers and the neural models utilized the statistic distribution of phonetic radicals in naming familiar characters. However, these effects can not be applied in unknown character naming since the speakers don't know the statistics of these characters. Therefore, we proposed a new metric (saliency of the phonetic radical) to measure how the phonetic radicals influence the speaker's unknown character naming behavior. Saliency is defined as the fraction of the \regular characters among all characters sharing the same phonetic radical. For example, the phonetic radical 青 $<$qing$>$ appeared in 12 characters in Table~\ref{tab:0}, among which 4 characters (清，情，圊，晴) are \regular. Thus the saliency score of 青 $<$qing$>$ is 0.33 (4/12). The more salient a phonetic radical is, the more likely the character that contains it is pronounced the same as its pinyin. 

We hypothesized that the human speakers would show a saliency effect in unknown character naming - they would name the characters more accurately if the phonetic radical is more salient. We expected to find a similar saliency effect in the models. In addition, we also closely examine the models' answers and humans' answers to investigate if the models can represent the human speaker's behavior.
\section{Data}
The base character dataset consists of 4,341 Chinese characters constructed from the \href{https://github.com/cjkvi/cjkvi-ids/blob/master/ids-analysis.txt}{IDS dataset} in CHISE project \cite{morioka2008chise}. The original IDS (Ideographic Description Sequence) dataset contains 18,347 characters used in China, Japan, and Korea with the decomposition of each character's phonetic and semantic radicals.\footnote{The phonetic and semantic radicals are decomposed according to \textit{Shuowen Jiezi} 《說文解字》.} The character selection criterion include: 1) is used in Chinese; 2) is a phono-semantic compound; 3) has a left-right structure\footnote{Following \citet{hsiao2004connectionist}, we only selected left-right structure to make sure that the character's structure is not a variable in our study.}. The character's pinyin, along with its phonetic and semantic radical's pinyin, was collected using the \href{https://pypi.org/project/pinyin/}{pinyin package}. The frequency of each character was extracted from BLCU Corpus Center \cite{xun2016}.
We further labeled each character's regularity: \regular, \alliterating, \rhyming, and \rad as described in Table \ref{tab:0}. In addition, we calculated each phonetic radical's saliency.

There are 660 radicals after decomposing the 4,341 characters, among which 46 radicals only serve as the semantic radicals; 493 radicals only serve as the phonetic radicals; 121 radicals serve as both semantic and phonetic radicals. Each radical appears in 7 characters on average, with a range of 1 to 30. 80\% of the characters in our database have the phonetic radical on the right, with many exceptions, e.g., the semantic radical `戈' $<$ge$>$ always appears on the right.

\subsection{Test Data}
\begin{figure}[!ht]
    \centering
    \includegraphics[width = \linewidth]{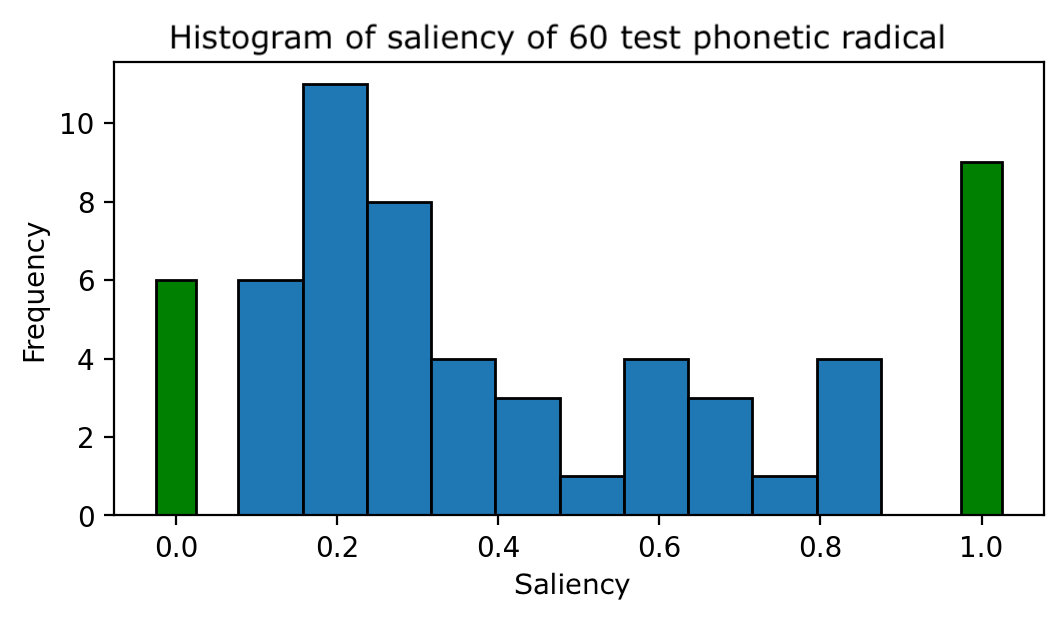}
    \caption{\small The histogram of saliency scores for 60 test characters' phonetic radicals.}
    \label{fig:test_cons}
\end{figure}

We selected 60 characters with different phonetic radicals from the dataset as our test data, which are listed in Table 14 in Appendix B. The test characters are selected following two criteria to ensure that human speakers are unfamiliar with the character, while familiar with the phonetic radicals, 1) the character appears less than 5 times in the whole corpus, 2) the phonetic radical in each character appears in more than 4 other characters.
The average saliency score for these phonetic radicals is 0.43, with the score distribution shown in Figure \ref{fig:test_cons}.
Among these characters, 22 of them have more than one pinyin, e.g., `硞' ($<$que$>$, $<$ke$>$, $<$ku$>$), which yields 88 pinyins for 60 characters. The distribution of the regularity type for the test characters is shown in Table~\ref{tab:new}.  

\begin{table}[!ht]
\small
\centering
\begin{tabular}{p{0.35\linewidth}p{0.23\linewidth}p{0.23\linewidth}}
\toprule
Regularity Type & \# pinyin & \# character \\
\midrule
\textit{regular} & 30 & 30 \\
\textit{alliterating} & 6 & 6 \\
\textit{rhyming} & 24 & 19 \\
\textit{irregular} & 28 & 20 \\
\bottomrule
\end{tabular}
\caption{\small The distribution of regularity types of pinyins and characters for the test data.}
\label{tab:new}
\end{table}

\subsection{Training Data}
\label{sec:train}
We exclude the 60 test characters and use the rest of the characters as our training data (4,281). The \regular is the most common type (42.7\%), followed by \rad, \rhyming, and \alliterating. Since many of the characters have extremely low frequency and are not known to the Chinese speakers, we used three training datasets with characters of different frequencies to represent the native speakers' vocabulary size. The \dall dataset used all 4,281 characters. The \dmid dataset consists of 2,140 characters whose frequencies are in the top 50\% percentile. The \dhigh dataset consists of 1,070 characters with frequencies in the top 25\% percentile. The statistics of these training sets are shown in Table \ref{tab:train}. Each training set has similar regularity distribution. 

\begin{table}[!t]
\small
\centering
\begin{tabular}{p{0.32\linewidth}p{0.15\linewidth}p{0.15\linewidth}p{0.15\linewidth}}
\toprule
Training Data & \dall & \dmid & \dhigh \\
\midrule
\# characters & 4,281 & 2,140 & 1,070 \\\midrule
\regular (\%) & 42.7 & 43.3 & 42.1 \\
\alliterating (\%) & 7.8 & 8.1 & 8.7 \\
\rhyming (\%)& 23.6 & 23.3 & 22.5 \\
\rad (\%)& 25.9 & 25.3 & 26.7 \\
\bottomrule
\end{tabular}
\caption{\small Number of characters and percentage of regularity types for our training datasets.}
\label{tab:train}
\end{table}

\section{Human Experiment}
\label{sec:human}
55 native speakers of Mandarin participated in this study. All of them are able to read and write in traditional Chinese scripts and pinyin. The average age is 26.3, and 80\% of them have an education background of college or above. In the experiments, they were asked if they knew the character and prompted to type the pinyin of the character. The detailed experiment procedure and sample questions are described in Appendix A. 

\subsection{Results: Human Answer's Accuracy} 
In general, the test characters are unknown to the participants.\footnote{Very few participants indicated that they know one or two test characters. For those who indicated that they knew the character, they still answered its pinyin incorrectly.} The accuracy is calculated on the syllable onset and final, ignoring the tone, since tones are more affected by the speaker's accent than syllable onsets and finals. For polyphone characters, as long as the participant named one correct pinyin, we counted it as correct. 
The average accuracy for all participants is 45.3\% (27 out of 60 characters), with a range of 26.7\% - 68.3\%. Some characters are more difficult to name than others. For example, 8 characters' accuracies are 0, meaning that none of the participants named them correctly. The character's accuracy is calculated as the proportion of participants who named it correctly, ranging from 0 - 98.2\%. There is a strong positive correlation between the character's accuracy and its phonetic radical's saliency (r = 0.62), which confirms our hypothesis about the saliency effect. The more salient the phonetic radical is, the more participants named the character correctly. 
The accuracy measures how well the human speakers can grasp the grapheme-phoneme distributional patterns in Chinese. The results show that even native speakers can not accurately predict the pronunciation of an unknown character, which reflects the complex nature of Chinese grapheme-phoneme mapping system. 
\subsection{Results: Human Answer's Variability} Since the participants named the character's pinyin differently, each character has a variety of unique answers. On average, each character has 6.7 answers, with a minimum of 2 answers and a maximum of 15 answers. The number of answers is negatively correlated with the saliency of the phonetic radical (r = -0.51), that the more salient the phonetic radical, the fewer number of answers the speakers guessed.  

We defined 5 answer types based on regularity. The participants either guessed the character's pinyin the same as its phonetic radical's (\regular), or changing the syllable final (\alliterating), the syllable onset (\rhyming), or both (\rad), or mistakenly used the semantic radical to name the character (\sem).\footnote{When examining the data, we found that some participants named the character the same as its semantic radical. We loosely defined this type of error as \sem type. It could also be that the participants applied \rad on the phonetic radical, and the pinyin happened to be the same as the semantic radical's pinyin. However, there's no way to confirm it. We asked some of our participants (with linguistic background) to explain how they guessed the pinyin, and none of them could articulate their thinking process.} %
We presented the answer types for character `煔' as an example in Table \ref{tab:shan}, and defined the production probability $P_{p}$ by the proportion of participants named that answer type.

The average production probability for each type is listed in Table \ref{tab:answer_type}. Most of the participants are able to identify the phonetic radical correctly, as the average production probability of the semantic type is only 2\%. The regular answer type has the highest production probability (58\%), suggesting that the participants are more likely to name the character the same as its phonetic radical. The production probabilities of answer types for each character are plotted in Figure \ref{fig:variety} in Section \ref{sec:overlap}.
\begin{table}[!ht]
\small
\centering
\begin{tabular}{p{0.25\linewidth}p{0.38\linewidth}p{0.15\linewidth}}
\toprule
\multicolumn{3}{l}{煔 $<$shan$>$, $<$qian$>$, `sparkle'}\\
\multicolumn{3}{l}{Phonetic Radical: 占 $<$zhan$>$, `to seize'}\\
\multicolumn{3}{l}{Semantic Radical: 炎 $<$yan$>$, `fire'}\\
\midrule
Answer Type & Answer(s) & $P_{p}$ (\%)\\
\midrule
\regular & $<$zhan$>$ & 36.4 \\
\alliterating & $<$zhen$>$ & 1.8 \\
\rhyming & $<$dan$>$ & 3.6 \\
\rad & \makecell[l]{$<$nian$>$, $<$jian$>$, \\$<$dian$>$,$<$pou$>$\\$<$yi$>$, $<$tie$>$} & 34.6 \\
\sem & $<$yan$>$ & 23.6 \\
\bottomrule
\end{tabular}
\caption{\small The answer types and production probability of human answers for polyphone `煔'.}
\label{tab:shan}
\end{table}

\begin{table}[!ht]
\small
\centering
\begin{tabular}{p{0.25\linewidth}p{0.3\linewidth}p{0.3\linewidth}}
\toprule
Answer Type & Average $P_{p}$ (\%) & Range of $P_{p}$ (\%) \\
 \midrule
\regular  & 58.0\std{25.8} & 0-98.2 \\
\alliterating  & 6.8\std{16.4} & 0-81.8 \\
\rhyming & 13.0\std{18.9} & 0-81.8 \\
\rad & 20.6\std{20.0} & 0-72.7 \\
\sem  & 1.6\std{4.6} & 0-23.6 \\
\bottomrule
\end{tabular}
\caption{\small The average production probability and its range for each answer type in human answers.}
\label{tab:answer_type}
\end{table}

\section{Transformer Model}
To model the joint probability of the syllable onset and final, we used seq-to-seq transformers \cite{vaswani2017attention} to generate the pinyin of Chinese characters trained from scratch.\footnote{We did not use a classification model because there are certain rules in pinyin formation (e.g., /ü/ can not follow /b/, /p/, /m/, /f/), which requires the model to learn the syllable onsets and finals jointly.}

\subsection{Experiment Setups}
Both encoder and decoder of all our models had 2 layers, 4 attention heads, 128 expected features in the input, and 256 as the dimension of the feed-forward network model. 
For training, we split the dataset into train/dev splits of 90-10, and replace those tokens that appear once in training data by \textlangle unk\textrangle. We also set dropout to 0.1, batch size to 16, and used Adam optimizer \cite{DBLP:journals/corr/KingmaB14} with varied learning rates in the training process computed according to \citet{vaswani2017attention}. We used 5 different random seeds, and trained 40 epochs with early stopping %
for all of our experiments. For inference, we set beam size to 3. 

\subsection{Experiment 1}
\label{sec:test_acc}

We trained a set of models to simulate the grapheme-phoneme mapping process in Chinese speakers. Our \base model used the phonetic radical's orthographic forms to generate syllable onset and final (without tone) of the target character.
We further examined whether identifying the phonetic radical before generating the syllable onset and final would improve the model's performance.  %
We labeled the phonetic radical's position (left or right) with two methods: \labm and \labs. \labm used the true position of the phonetic radical as the ground truth label. 
Besides, since human speakers do not always identify the phonetic radical's position correctly,
\labs labeled the position of the phonetic radical based on the phonetic similarity. We calculated the phonetic similarity between the character's pinyin and the two radicals' pinyins using the Chinese Phonetic Similarity Estimator \cite{li2018dimsim}. The radical with higher phonetic similarity was labeled as the phonetic radical.\footnote{For example, the character `烙' $<$luo4$>$ (`flatiron') consists of the semantic radical `火' $<$huo3$>$ (`fire') and the phonetic radical `各' $<$ge4$>$ (`each'). The distance between $<$luo4$>$ and $<$huo3$>$ is 7.5, and the distance between $<$luo4$>$ and $<$ge4$>$ is 35.6. For \labs, the output radical should be `left', although the left radical `火' is the semantic radical.} We further labeled the regularity type of the characters based on \labm and \labs, hence yielding \labmr and \labsr. Examples of input and gold output in the training data are shown in Table~\ref{input_example}. All the models were trained on \dall, \dmid, and \dhigh datasets as described in section~\ref{sec:train}.

Since previous studies suggested that the regularity and consistency effects are more prominent for the characters with low frequency than high frequency  \cite[e.g.,][]{ziegler2000phonology,chen2009homophone}, the frequency of the known characters might also influence how participants predict the unknown characters. We further added the frequency label as an input feature in the full training data as the \dallf model. The characters  were categorized into four categories based on their frequency: `rare' (frequency = 1), `low' (1 < frequency $\leq$ 50\% percentile), `mid' (50\% percentile < frequency $\leq$ 75\% percentile) and `high' (frequency > 75\% percentile).%
The distribution of regularity types is similar for the characters with different frequencies. The summary of the number of characters and each regularity type can be found in Appendix B, Table~\ref{app_tab:3_freq}. 

 In addition, we added two conditions for output in training all models: Shuffling and Adding tones. We shuffle the position of the syllable onset and final in model output to explore the impact of the generated order since we don't know if the human speakers identify the syllable onset or syllable final first in character naming. We also add tones before the `End' token in the generation to see whether it improves the model performance. Examples of input and output of the conditions are shown in Table \ref{input_example}. In total, there are 80 types of models with different settings. 

\begin{table}[t!]
\small
\centering
\begin{tabular}{p{0.18\linewidth}p{0.68\linewidth}}\toprule
Input & Begin, 火, 各, End \\
\midrule
Model & Output \\
\midrule
\base & Begin, l, uo, End \\
\labm & Begin, right, l, uo, End \\
\labs & Begin, left, l, uo, End \\
\labmr & Begin, right, irregular, l, uo, End \\
\labsr & Begin, left, rhyming, l, uo, End \\
\midrule\midrule
Condition & Input \\\midrule
\dallf & Begin, 火, 各, high, End \\
\midrule
Condition & Output (\base model as an example) \\
\midrule
{[+}Shuffle{]} & Begin, uo, l, End \\
{[+}Tone{]} & Begin, l, uo, 4, End \\
\bottomrule
\end{tabular}
\caption{\small \label{input_example}Input and gold output in the training data of our models and conditions for character `烙'$<$luo4$>$, tokens are separated by comma.}
\end{table}

\paragraph{Accuracy Results}
We calculated the test accuracy the same way as for the human data: we only counted the accuracy of the syllable onset and final. For polyphone characters, as long as the model predicted one correct pinyin, it is counted as correct. The average accuracy of all 400 models (80 types x 5 random seeds) is 42.1\%, which is significantly lower than the human's accuracy (45.3\%, t = 3.15, p<0.01). The average accuracy of each type of model is listed in Table~\ref{tab:3_dfreq}. The best performing model is \dallf with \labm without tone and with shuffling, which achieved an accuracy of 50.3\%. Compared to the \base model, adding the label of phonetic position label and the character's regularity label usually could improve the model's accuracy. Adding tone would generally hurt the model's accuracy. Shuffling the syllable onset and final and adding the frequency label in the input would not change the model's accuracy.

\begin{table}[!ht]
\small
\centering
\begin{tabular}{llllll}
\toprule
data & label & -T-S & -T+S & +T-S & +T+S \\
\midrule
\multirow{5}{*}{\dall} & \base & 49.3 & 49.3 & 42.3 & 46.0 \\
 & \labm & 48.0 & 49.7 & 45.3 & 47.7 \\
 & \labs & 46.0 & 45.3 & 42.3 & 48.7 \\
 & \labmr & 47.0 & 48.7 & 48.7 & 49.7\\
 & \labsr & 44.0 & 47.3 & 45.0 & 48.3 \\
 \midrule

\multirow{5}{*}{\dmid} & \base & 41.7 & 41.3 & 38.7 & 41.7 \\
 & \labm & 44.3 & 43.0 & 44.0 & 42.3 \\
 & \labs & 41.3 & 43.3 & 41.3 & 42.3 \\
 & \labmr & 42.0 & 40.3 & 39.7 & 44.3 \\
 & \labsr & 37.7 & 42.7 & 39.0 & 42.0 \\
 \midrule

\multirow{5}{*}{\dhigh} & \base & 28.7 & 32.3 & 29.3 & 32.3 \\
 & \labm & 36.3 & 34.7 & 30.7 & 35.3 \\
 & \labs & 32.7 & 36.0 & 30.0 & 34.0 \\
 & \labmr & 31.3 & 31.7 & 31.3 & 32.0 \\
 & \labsr & 32.3 & 32.0 & 31.0 & 33.7 \\
 \midrule

\multirow{5}{*}{\begin{tabular}[c]{@{}l@{}}\textsc{all+}\\ \textsc{freq}\end{tabular}} & \base & 46.7 & 47.0 & 47.7 & 46.7 \\
 & \labm & 49.7 & 50.3 & 47.3 & 47.0 \\
 & \labs & 45.3 & 47.3 & 47.0 & 48.3 \\
 & \labmr & 46.3 & 49.3 & 47.0 & 48.0 \\
 & \labsr & 47.7 & 44.7 & 44.0 & 47.7 \\
 \bottomrule
\end{tabular}
\caption{\label{tab:3_dfreq}The average accuracy (over 5 seeds) on test set for models trained on \dhigh, \dmid, or adding frequency label as input features on \dall. {+}T, {-}T, {+}S, {-}S refers to adding tone, no tone, shuffling, and no shuffling, respectively.}
\end{table}

\subsection{Experiment 2}
In Experiment 1, the input of our models only used the orthographic form of the radicals, which is how the previous literature described the Chinese grapheme-phoneme mapping process. However, the models might not have enough data to learn the full mapping from radicals to pinyin because many radicals only appeared once or twice in the training data since we only included compound characters with the left-right structure. For example, the phonetic radical `乘' $<$cheng$>$ only occurred once in the character `剩' $<$sheng$>$ in the training data.\footnote{We choose the first pinyin from the pinyin package for polyphone radicals.} The models would not be able to accurately learn the pinyins of these radicals. However, human speakers know the pinyin of most radicals, since many radicals are also commonly used as stand-alone characters, e.g., `乘' is a stand-alone character meaning `to multiply'. In order to better model the human speakers, it is necessary to inject pinyin of the radicals as external information to the model. The model would also benefit from the added radicals' pinyin to generate the character's pinyin. 

In addition, pinyin also plays an important role in modern Chinese speakers' reading and spelling experience. Pinyin is a Romanized phonetic coding system created in 1958 to promote literacy \cite{zhou1958}. In the information age, pinyin has become indispensable in Chinese speakers' lives because it's the dominant typing system for computers, smart phones, and electronic devices. The prevalent experience of typing characters through pinyin has challenged the traditional view that Chinese characters are processed purely through orthographic forms \cite{tan2013china}. Many recent studies have found that pinyin mediates the character recognition process\cite{chen2017effect,lyu2021comparison,yuan2022role}. To better capture modern Chinese speakers' character naming process, it is necessary to incorporate the radical's orthographic form as well as its pinyin in our models. 

Therefore, in Experiment 2, we added the radical's pinyin (syllable onset, syllable final, and tone) in the input, as shown in Table~\ref{tab:EXP2}. We used the same model variations as in Experiment 1\footnote{For the output, we added \labm, \labs, \labmr, \labsr as well as adding tone and shuffling. For the input, we added frequency label to create \dallf.} and trained 80 different types of models (5 random seeds for each type) with the new input. The training settings are the same as Experiment 1. 
\begin{table}[!ht]
    \centering
    \small
    \begin{tabular}{ll}
    \toprule
       Input  & Begin, 火, h, uo, 3, End, 各, g, e, 4, End \\
    \bottomrule
    \end{tabular}
    \caption{Input in the training data for Experiment 2 using `烙' $<$luo4$>$ as an example.}
    \label{tab:EXP2}
\end{table}

\paragraph{Accuracy Results}
Adding pinyin to the input has increased the model's accuracy.\footnote{We can not fully rule out the possibility that the increased accuracy is due to the model having longer inputs with pinyin instead of the model making use of the phonetic information. However, the input length might not have a significant impact on the models because our models with frequency labels (\dall vs \dallf) also vary in input lengths but the accuracies didn't change much.} The average accuracy of 400 models in Experiment 2 is 47.4\%, which is significantly higher than the human's accuracy (t = -2.7, p <0.01). The accuracy for each type of model is listed in Table~\ref{exp2_tab:3_dfreq} in Appendix B. The best performing model is \dallf with \labmr without tone and with shuffling, which achieved an accuracy of 55\%. The effects of different labels, adding tone, and shuffling are similar to the models in Experiment 1. 
\section{Comparison Between Models' Results and Human Behaviors}
\label{sec:4.3}
In this section, we compared transformer models' results in Experiments 1 (\MODA) and 2 
(\MODB) with human performance. Since human participants are different, i.e., they have different vocabularies, and they may use different strategies to identify the phonetic radical, we used all 80 models in each experiment to represent the human variety. Following \citet{corkery2019we}, each random initialization was also treated as an individual participant. Therefore, the sample size for the human participants is 55, and the sample size for the models in each experiment is 400 (80 models $\times$ 5 initialization). We focused on three types of similarities: 1) accuracy, i.e., do humans and models show similar accuracy on each character? 2) overlap, i.e., do humans and models predict the same pinyin for each character? 3) variability, i.e., do humans and models have similar answer regularity patterns?

\paragraph{Accuracy}
We calculated each character's accuracy for \MODA and \MODB. First, both models showed saliency effect: the model's character accuracy is positively correlated with saliency score (Pearson $r$ = 0.48 for \MODA and $r$ = 0.57 for \MODB), which is not significantly different from human's saliency correlation ($r$ = 0.62). In addition, there's a strong correlation between human character accuracy and both models' character accuracy (\MODA $r$ = 0.79, \MODB $r $= 0.88), suggesting that the humans and models are in high agreement. In conclusion, the transformer models' answers are very similar to the human answers in terms of character accuracy.

\begin{figure}[t!]
    \centering
    \includegraphics[width = 0.5\textwidth]{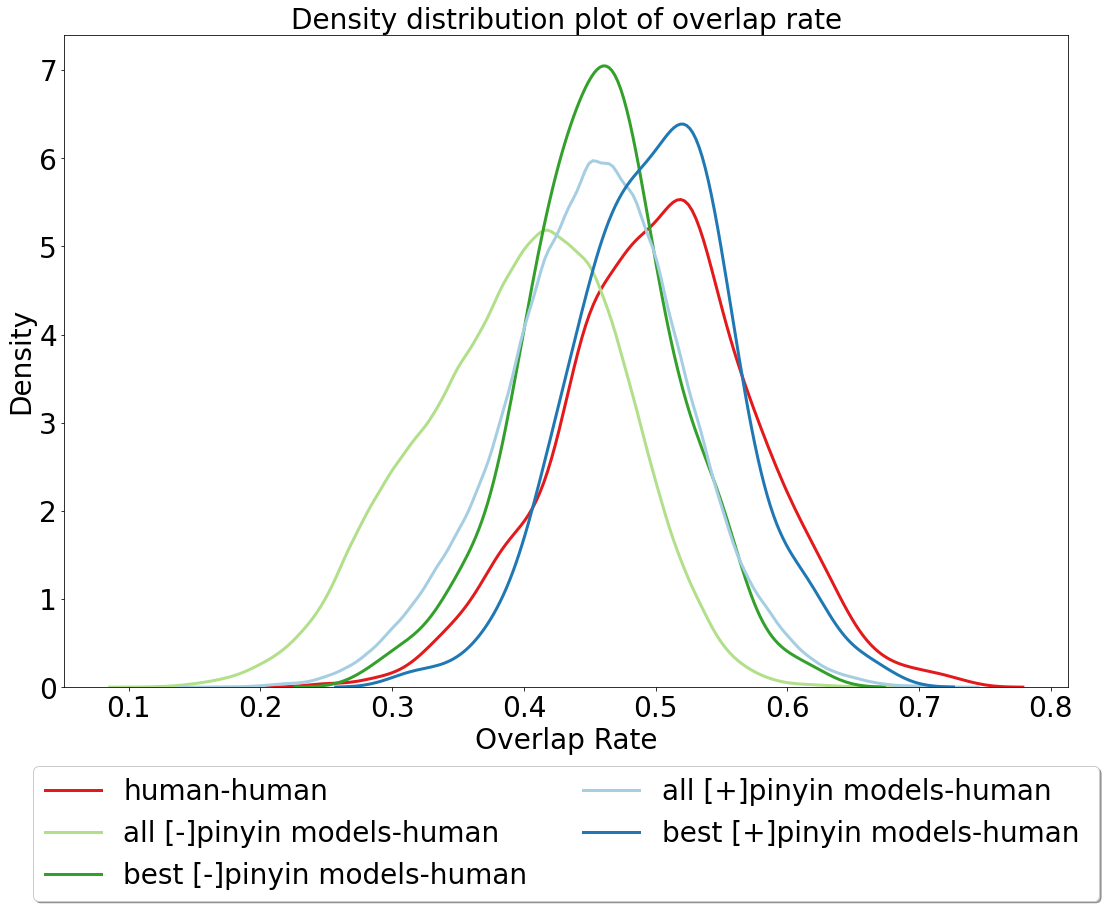}
    \caption{\small Density plot of the overlap rate.}
    \label{fig:overlap}
\end{figure}

\paragraph{Overlap}
\label{sec:overlap}
The overlap rate was computed to measure to what extent different human speakers (and models) predict the same answers for each character. For example, if participant 1 and 2 have 30 same answers, then the overlap rate = 50\% (30/60). Among 1,485 answer pairs of 55 human speakers, the average overlap rate of human-human is 50.2\%, with a range of 25.0\% - 73.3\%. For \MODA, among 400 models and 55 speakers, the average overlap rate for 22,000 answer pairs is 39.6\%, with a range of 12\% - 66.7\%. For \MODB, the average overlap rate of 22,000 answer pairs is 45.2\%, with a range of 16.7\% - 71.7\%. Both models' overlap rates are significantly lower than the human-human overlap rate, and \MODB's overlap rate is significantly higher than \MODA.  In addition, we computed the human-model overlap rate for different models, with 275 answer pairs for each model (5 random seeds $\times$ 55 human speakers).\footnote{See the detailed overlap results for \MODA and \MODB in Table~\ref{app_tab: overlap}, Appendix B.}
The best model for \MODA is \dall with \labmr with tone and without shuffling, with an overlap rate of 45.6\%. 
The best model for \MODB is \dallf with \labs with tone and without shuffling, with an overlap rate of 50.1\%, which is not significantly different from human-human overlap rate. The overlap results are summarized in Table \ref{tab:overlap}. The density plot of the overlap rate for human-human, human-all models, and human-best model is shown in Figure \ref{fig:overlap}. In general, the humans' answers are more similar to each other than to the models' answers. \MODB's answers are more similar to human answers than \MODA. 

\begin{table}[!ht]
\small
\centering
\begin{tabular}{lll}
\toprule
 & Overlap rate& Range \\
 \midrule
Human - Human & 50.2\std{7.0} & 25.0-73.3 \\
\midrule
\multicolumn{3}{l}{Transformer models - Human}\\
\midrule
All \MODA & 39.6*\std{7.6} & 11.7-66.7\\
Best \MODA & 45.6*\std{5.8} & 28.3-61.7\\
All \MODB & 45.3*\std{7.0} & 16.7-71.7 \\
Best \MODB & 50.1\std{6.1} & 31.7-66.7 \\
\bottomrule
\multicolumn{3}{l}{\scriptsize*indicates significantly smaller than 50.2}
\end{tabular}
\caption{\small The average overlap rate (\%) and its range for human-human and transformer-human comparison.}
\label{tab:overlap}
\end{table}

\paragraph{Variability}
\begin{figure*}[!ht]
    \centering
    \includegraphics[width=0.95\textwidth]{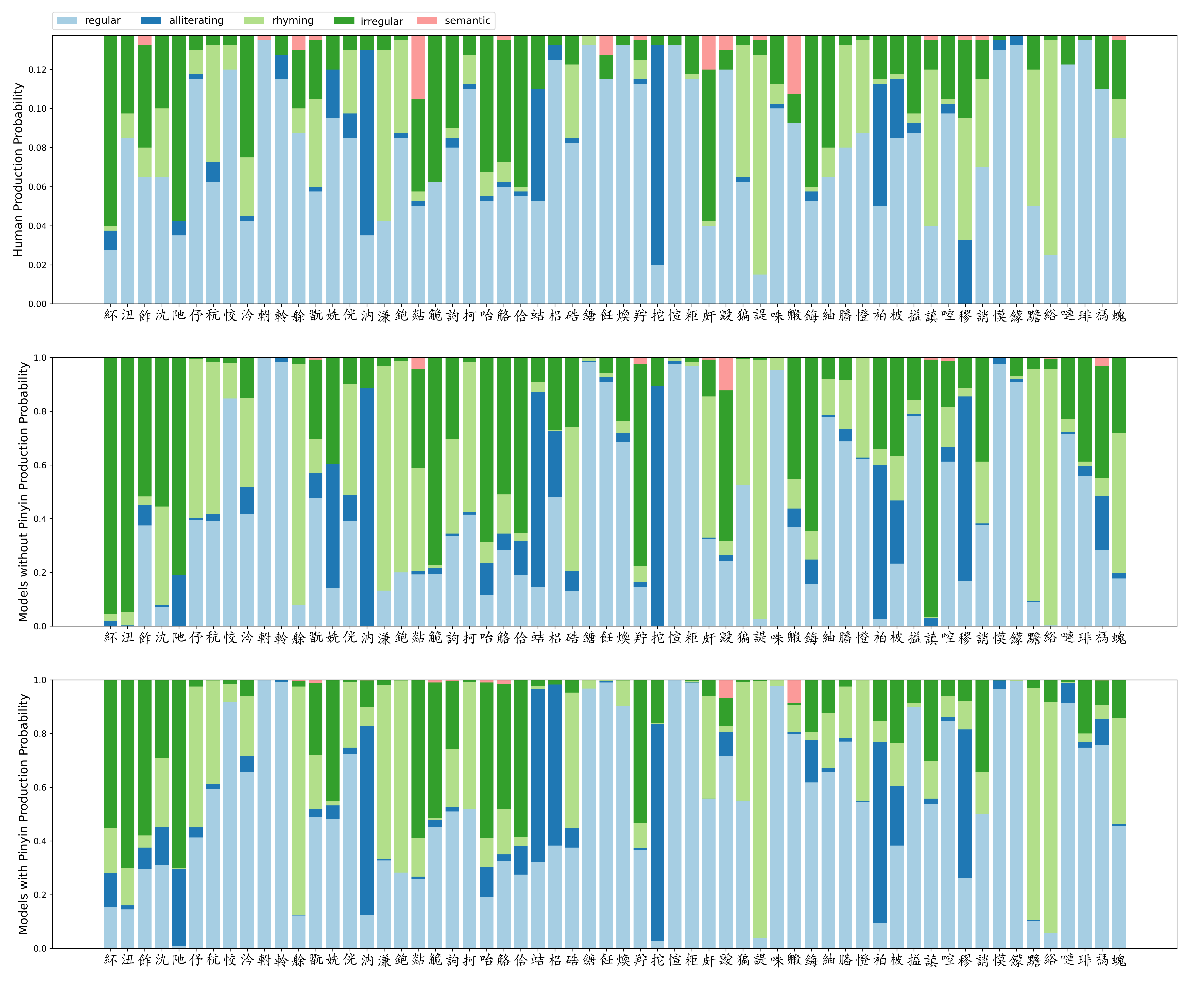}
    \caption{\small The production probability of 5 answer types produced by humans (top), \MODA (middle), and \MODB (bottom).}
    \label{fig:variety}
\end{figure*}
Like human speakers, transformer models also produce different answers for each character. We categorized these answers based on their regularity type and calculated the models' averaged production probability ($P_p$) for each answer type, as listed in Table~\ref{tab:Ttype}. We further calculated Spearman correlation ($\rho$) and Pearson correlation ($r$) between the production probability of each type in human answers and the models' answers on each character (N = 60). All the regularity types are highly correlated except for the \textit{semantic} type. The models did not produce as many \textit{semantic} type answers as humans, suggesting the models are better at identifying the phonetic radical than humans. In addition, we also calculated the cross-entropy between the humans and the models on the production probability of 5 regularity types. The cross-entropy for \MODA is $\mathrm{H}$(human, \MODA) = 1.79 and for \MODB is $\mathrm{H}$(human, \MODB) = 1.74, suggesting that \MODB is slightly more similar to the human than the \MODA. 

\begin{table}[!ht]
\small
\centering
\begin{tabular}{lllll}
\toprule
 & \makecell{\textsc{model}\\\scriptsize\textsc{[-pinyin]}} & Cor. & \makecell{\textsc{model}\\\scriptsize\textsc{[+pinyin]}} & Cor. \\
 \midrule
\multirow{2}{*}{\textit{Reg.}} & \multirow{2}{*}{39.4*\std{32.6}} & $\rho$ 0.72 & \multirow{2}{*}{52.9\std{30.3}} & $\rho$ 0.70 \\
 &  & $r$ 0.71 &  & $r$ 0.72 \\
 \hline
\multirow{2}{*}{\textit{Alli.}} & \multirow{2}{*}{10.9\std{21.4}} & $\rho$ 0.59 & \multirow{2}{*}{10.1\std{19.7}} & $\rho$ 0.51 \\
 &  & $r$ 0.95 &  & $r$ 0.85 \\
\hline
\multirow{2}{*}{\textit{Rhym.}} & \multirow{2}{*}{22.6*\std{28.4}} & $\rho$ 0.64 & \multirow{2}{*}{20.6\std{25.1}} & $\rho$ 0.70 \\
 &  & $r$ 0.58 &  & $r$ 0.62 \\
 \hline
\multirow{2}{*}{\textit{Irr.}}& \multirow{2}{*}{26.6\std{28.1}} & $\rho$ 0.55 & \multirow{2}{*}{16.1\std{20.6}} & $\rho$ 0.67 \\
 &  & $r$ 0.50 &  & $r$ 0.58 \\
 \hline
\multirow{2}{*}{\textit{Sem.}}& \multirow{2}{*}{0.5*\std{1.7}} & $\rho$ 0.39 & \multirow{2}{*}{0.2*\std{0.9}} & $\rho$ NA$\dagger$ \\
 &  & $r$ 0.30 &  & $r$ 0.09 \\
\bottomrule
\multicolumn{5}{l}{\scriptsize {\begin{tabular}[c]{@{}l@{}}* indicates significantly different from human's  ($P_p$).\\NA$\dagger$ is due to too many zeros in the data that the correlation\\ cannot be calculated\end{tabular}}}
\end{tabular}
\caption{\small The average production probability ($P_p$) of each answer type and their correlation ($\rho$ and $r$) with humans for \MODA and \MODB.}
\label{tab:Ttype}
\end{table}

The production probability of different regularity types for each character is shown in Figure \ref{fig:variety}. The answer type patterns are very similar for humans and models except for the \textit{semantic} type. Humans produced \textit{semantic} type answers for 15 characters, while both our models produced \textit{semantic} type for fewer characters with a much smaller production probability. This implied that phonetic radicals are identified differently by humans and transformer models. Humans are affected by a wide range of linguistic knowledge in identifying the phonetic radical, including the semantic meaning of the radical, vocabulary size, and reading comprehension \cite{anderson2013learning,yeh2017lexical}. The models did not receive these extra inputs, and thus did not closely capture human behavior on \textit{semantic} answer type.

\section{Conclusion and Discussion}
\paragraph{Conclusion} We evaluated transformer models and human behaviors on an unknown Chinese naming task. This task is difficult for both humans and transformer models, as the average accuracy is lower than 50\%. Humans have higher accuracy than \MODA and lower accuracy than \MODB, and the models and the humans have very similar performances. First, saliency effects were found in both human data and the models' results, suggesting that both models and humans utilize the statistical distribution of the phonetic radical to infer the character's pinyin. Further, although humans' answers are more similar to each other, our models also achieved a substantial overlap with humans' answers. Additionally, the production probability of each answer type is highly correlated between models and humans (except for \textit{semantic} type), suggesting that both models and humans are able to apply all regularity patterns in producing answers. Finally, models with radical's pinyins in the input are more similar to humans and achieved higher accuracy. 

\paragraph{Capturing quasi-regularity}
Our work is also related to the long-standing criticism that the neural networks may only learn the most \textit{frequent} class and can not extend other minority classes, thus would fail to learn the quasi-regularity in languages \cite{marcus1995german}. Previous studies on morphological inflections have shown that the neural models overgeneralized the most frequent inflections on nonce words and had almost no correlation with human's production probability on the less frequent inflections (e.g., $\rho$ = 0.05 for the /-er/ suffix in German plural \cite{mccurdy2020inflecting}, and $r$ = 0.17 for irregular English verbs \cite{corkery2019we}). However, our results showed that the transformer models could learn the quasi-regularity in Chinese character naming, that the models produce all answer types, and the production probability of each type is highly correlated with human data. 

However, our results do not contradict the previous studies. Chinese character naming and morphological inflection both exhibit quasi-regularity, but the two domains are very different: the patterns in Chinese character naming are less rule-governed. This paper's contribution to the debate of quasi-regularity in language processing is not to provide a `yes' or `no' answer; instead, we used a novel task and showed that the neural models have the potential to model human behaviors in learning quasi-regularity. We hope our study could inspire future work in this field to apply diverse tasks and conduct more detailed examinations of neural models' ability in learning quasi-regularity. 

\paragraph{Modeling Chinese reading with neural network} Our study also contributed to the current debate of whether reading skill is acquired by domain-general statistical learning mechanism \cite{plaut2005connectionist}, or language-specific knowledge such as DRC model \cite{coltheart2001drc}. Our results demonstrated that a general statistical learning mechanism (implemented as the transformer model) could learn the Chinese grapheme-phoneme mapping. We not only successfully simulated the general saliency effects in human's unknown character naming behavior, but also showed in details that the answers produced by models and humans are highly similar. 
Another contribution to modeling Chinese reading is that we are the first study that incorporated the radicals' pinyin in the model. Models with pinyin as input not only had better accuracy, but also are more similar to human behavior. Our results echoed the recent literature on the pinyin effect. For modern Chinese speakers who type characters through pinyin more often than hand-writing characters, pinyin can be an important mediator for the grapheme-phoneme mapping process.

\section{Acknowledgements}
We would like to thank the anonymous reviewers and action editor Micha Elsner for their valuable feedback. We would also like to thank Kevin Gimpel, Allyson Ettinger, Virginia Valian, Martin Chodorow, and Kyle Gorman for their insightful discussions and suggestions.

\bibliography{anthology,custom}

\begin{thebibliography}{39}
\expandafter\ifx\csname natexlab\endcsname\relax\def\natexlab#1{#1}\fi

\bibitem[{Anderson et~al.(2013)Anderson, Ku, Li, Chen, Wu, and
  Shu}]{anderson2013learning}
Richard~C Anderson, Yu-Min Ku, Wenling Li, Xi~Chen, Xinchun Wu, and Hua Shu.
  2013.
\newblock Learning to see the patterns in chinese characters.
\newblock \emph{Scientific Studies of Reading}, 17(1):41--56.

\bibitem[{Chen et~al.(2009)Chen, Vaid, and Wu}]{chen2009homophone}
Hsin-Chin Chen, Jyotsna Vaid, and Jei-Tun Wu. 2009.
\newblock Homophone density and phonological frequency in chinese word
  recognition.
\newblock \emph{Language and Cognitive Processes}, 24(7-8):967--982.

\bibitem[{Chen et~al.(2017)Chen, Luo, and Liu}]{chen2017effect}
Jingjun Chen, Rong Luo, and Huashan Liu. 2017.
\newblock The effect of pinyin input experience on the link between semantic
  and phonology of chinese character in digital writing.
\newblock \emph{Journal of Psycholinguistic Research}, 46(4):923--934.

\bibitem[{Chen(1996)}]{chen1996functional}
Yi-Ping Chen. 1996.
\newblock What are the functional orthographic units in chinese word
  recognition: The stroke or the stroke pattern?
\newblock \emph{The Quarterly Journal of Experimental Psychology: Section A},
  49(4):1024--1043.

\bibitem[{Coltheart(1978)}]{coltheart1978lexical}
Max Coltheart. 1978.
\newblock Lexical access in simple reading tasks.
\newblock \emph{Strategies of information processing}, pages 151--216.

\bibitem[{Coltheart et~al.(2001)Coltheart, Rastle, Perry, Langdon, and
  Ziegler}]{coltheart2001drc}
Max Coltheart, Kathleen Rastle, Conrad Perry, Robyn Langdon, and Johannes
  Ziegler. 2001.
\newblock Drc: a dual route cascaded model of visual word recognition and
  reading aloud.
\newblock \emph{Psychological review}, 108(1):204.

\bibitem[{Corkery et~al.(2019)Corkery, Matusevych, and
  Goldwater}]{corkery2019we}
Maria Corkery, Yevgen Matusevych, and Sharon Goldwater. 2019.
\newblock Are we there yet? encoder-decoder neural networks as cognitive models
  of english past tense inflection.
\newblock In \emph{57th Annual Meeting of the Association for Computational
  Linguistics}, pages 3868--3877. Association for Computational Linguistics
  (ACL).

\bibitem[{Ding et~al.(2004)Ding, Peng, and Taft}]{ding2004nature}
Guosheng Ding, Danling Peng, and Marcus Taft. 2004.
\newblock The nature of the mental representation of radicals in chinese: a
  priming study.
\newblock \emph{Journal of Experimental Psychology: Learning, Memory, and
  Cognition}, 30(2):530.

\bibitem[{Fang et~al.(1986)Fang, Horng, and Tzeng}]{fang1986consistency}
Sheng-Ping Fang, Ruey-Yun Horng, and Ovid~JL Tzeng. 1986.
\newblock Consistency effects in the chinese character and pseudo-character
  naming tasks.
\newblock \emph{Linguistics, psychology, and the Chinese language}, pages
  11--21.

\bibitem[{Glushko(1979)}]{glushko1979organization}
Robert~J Glushko. 1979.
\newblock The organization and activation of orthographic knowledge in reading
  aloud.
\newblock \emph{Journal of experimental psychology: Human perception and
  performance}, 5(4):674.

\bibitem[{Hsiao and Shillcock(2004)}]{hsiao2004connectionist}
Janet Hui-wen Hsiao and Richard Shillcock. 2004.
\newblock Connectionist modeling of chinese character pronunciation based on
  foveal splitting.
\newblock In \emph{Proceedings of the annual meeting of the cognitive science
  society}, volume~26.

\bibitem[{Hsiao and Shillcock(2005)}]{hsiao2005differences}
Janet Hui-wen Hsiao and Richard Shillcock. 2005.
\newblock Differences of split and non-split architectures emerged from
  modelling chinese character pronunciation.
\newblock In \emph{Proceedings of the Annual Meeting of the Cognitive Science
  Society}, volume~27.

\bibitem[{Hsiao and Shillcock(2006)}]{hsiao2006analysis}
Janet Hui-wen Hsiao and Richard Shillcock. 2006.
\newblock Analysis of a chinese phonetic compound database: Implications for
  orthographic processing.
\newblock \emph{Journal of psycholinguistic research}, 35(5):405--426.

\bibitem[{Hsu et~al.(2009)Hsu, Tsai, Lee, and Tzeng}]{hsu2009orthographic}
Chun-Hsien Hsu, Jie-Li Tsai, Chia-Ying Lee, and Ovid J-L Tzeng. 2009.
\newblock Orthographic combinability and phonological consistency effects in
  reading chinese phonograms: an event-related potential study.
\newblock \emph{Brain and Language}, 108(1):56--66.

\bibitem[{Hue(1992)}]{hue1992recognition}
Chih-Wei Hue. 1992.
\newblock Recognition processes in character naming.
\newblock In \emph{Advances in psychology}, volume~90, pages 93--107. Elsevier.

\bibitem[{Kingma and Ba(2015)}]{DBLP:journals/corr/KingmaB14}
Diederik~P. Kingma and Jimmy Ba. 2015.
\newblock \href {http://arxiv.org/abs/1412.6980} {Adam: {A} method for
  stochastic optimization}.
\newblock In \emph{3rd International Conference on Learning Representations,
  {ICLR} 2015, San Diego, CA, USA, May 7-9, 2015, Conference Track
  Proceedings}.

\bibitem[{Lam(2008)}]{lam2008exploratory}
Ho~Cheong Lam. 2008.
\newblock An exploratory study of the various ways that children read and write
  unknown chinese characters.
\newblock \emph{Journal of Basic Education}, 17(1).

\bibitem[{Lam(2014)}]{lam2014elaborating}
Ho~Cheong Lam. 2014.
\newblock Elaborating the concepts of part and whole in variation theory: The
  case of learning chinese characters.
\newblock \emph{Scandinavian Journal of Educational Research}, 58(3):337--360.

\bibitem[{Li et~al.(2018)Li, Danilevsky, Noeman, and Li}]{li2018dimsim}
Min Li, Marina Danilevsky, Sara Noeman, and Yunyao Li. 2018.
\newblock Dimsim: an accurate chinese phonetic similarity algorithm based on
  learned high dimensional encoding.
\newblock In \emph{Proceedings of the 22nd Conference on Computational Natural
  Language Learning}, pages 444--453.

\bibitem[{Li and Kang(1993)}]{li1993analysis}
Y~Li and JS~Kang. 1993.
\newblock Analysis of phonetics of the ideophonetic characters in modern
  chinese.
\newblock \emph{Information analysis of usage of characters in modern Chinese},
  pages 84--98.

\bibitem[{Lien(1985)}]{lien1985consistency}
Yunn-Wen Lien. 1985.
\newblock Consistency of the phonetic clues in the chinese phonograms and their
  naming latencies.
\newblock \emph{Psychological Department. National Taiwan University, Taipei}.

\bibitem[{Liu et~al.(2003)Liu, Chen, and Sue}]{liu2003regularity}
In-Mao Liu, SC~Chen, and IR~Sue. 2003.
\newblock Regularity and consistency effects in chinese character naming.
\newblock \emph{Chinese Journal of Psychology}, 45(1):29--46.

\bibitem[{Lyu et~al.(2021)Lyu, Lai, Lin, and Gong}]{lyu2021comparison}
Boning Lyu, Chun Lai, Chin-Hsi Lin, and Yang Gong. 2021.
\newblock Comparison studies of typing and handwriting in chinese language
  learning: a synthetic review.
\newblock \emph{International Journal of Educational Research}, 106:101740.

\bibitem[{Marcus et~al.(1995)Marcus, Brinkmann, Clahsen, Wiese, and
  Pinker}]{marcus1995german}
Gary~F Marcus, Ursula Brinkmann, Harald Clahsen, Richard Wiese, and Steven
  Pinker. 1995.
\newblock German inflection: The exception that proves the rule.
\newblock \emph{Cognitive psychology}, 29(3):189--256.

\bibitem[{McCurdy et~al.(2020)McCurdy, Goldwater, and
  Lopez}]{mccurdy2020inflecting}
Kate McCurdy, Sharon Goldwater, and Adam Lopez. 2020.
\newblock Inflecting when there's no majority: limitations of encoder-decoder
  neural networks as cognitive models for german plurals.
\newblock \emph{arXiv preprint arXiv:2005.08826}.

\bibitem[{Morioka(2008)}]{morioka2008chise}
Tomohiko Morioka. 2008.
\newblock Chise: Character processing based on character ontology.
\newblock In \emph{International Conference on Large-Scale Knowledge
  Resources}, pages 148--162. Springer.

\bibitem[{Plaut(2005)}]{plaut2005connectionist}
David~C Plaut. 2005.
\newblock Connectionist approaches to reading.
\newblock \emph{The science of reading: A handbook}, pages 24--38.

\bibitem[{Plaut et~al.(1996)Plaut, McClelland, Seidenberg, and
  Patterson}]{plaut1996understanding}
David~C Plaut, James~L McClelland, Mark~S Seidenberg, and Karalyn Patterson.
  1996.
\newblock Understanding normal and impaired word reading: computational
  principles in quasi-regular domains.
\newblock \emph{Psychological review}, 103(1):56.

\bibitem[{Seidenberg and McClelland(1989)}]{seidenberg1989distributed}
Mark~S Seidenberg and James~L McClelland. 1989.
\newblock A distributed, developmental model of word recognition and naming.
\newblock \emph{Psychological review}, 96(4):523.

\bibitem[{Tan et~al.(2013)Tan, Xu, Chang, and Siok}]{tan2013china}
Li~Hai Tan, Min Xu, Chun~Qi Chang, and Wai~Ting Siok. 2013.
\newblock China’s language input system in the digital age affects
  children’s reading development.
\newblock \emph{Proceedings of the National Academy of Sciences},
  110(3):1119--1123.

\bibitem[{Tsai et~al.(2005)Tsai, Su, Tzeng12, and Hung12}]{tsai2005consistency}
Jie-Li Tsai, Erica Chung-I Su, Ovid~JL Tzeng12, and Daisy~L Hung12. 2005.
\newblock Consistency, regularity, and frequency effects in naming chinese
  characters.
\newblock \emph{Language and Linguistics}, 6:75--107.

\bibitem[{Vaswani et~al.(2017)Vaswani, Shazeer, Parmar, Uszkoreit, Jones,
  Gomez, Kaiser, and Polosukhin}]{vaswani2017attention}
Ashish Vaswani, Noam Shazeer, Niki Parmar, Jakob Uszkoreit, Llion Jones,
  Aidan~N Gomez, {\L}ukasz Kaiser, and Illia Polosukhin. 2017.
\newblock Attention is all you need.
\newblock \emph{Advances in neural information processing systems}, 30.

\bibitem[{Xun et~al.(2016)Xun, Rou, Xiao, and Zhang}]{xun2016}
Endong Xun, Gaoqi Rou, Xiaoyue Xiao, and Jiaojiao Zhang. 2016.
\newblock 大數據背景下bcc語料庫單研製.[the construction of the bcc
  corpus in the age of big data].
\newblock \emph{語料庫語言學[Corpus Linguistics]}, 3(1):93--118.

\bibitem[{Yang et~al.(2009)Yang, McCandliss, Shu, and
  Zevin}]{yang2009simulating}
Jianfeng Yang, Bruce~D McCandliss, Hua Shu, and Jason~D Zevin. 2009.
\newblock Simulating language-specific and language-general effects in a
  statistical learning model of chinese reading.
\newblock \emph{Journal of Memory and Language}, 61(2):238--257.

\bibitem[{Yeh et~al.(2017)Yeh, Chou, and Ho}]{yeh2017lexical}
Su-Ling Yeh, Wei-Lun Chou, and Pokuan Ho. 2017.
\newblock Lexical processing of chinese sub-character components: Semantic
  activation of phonetic radicals as revealed by the stroop effect.
\newblock \emph{Scientific reports}, 7(1):1--12.

\bibitem[{Yuan et~al.(2022)Yuan, Segers, and Verhoeven}]{yuan2022role}
Han Yuan, Eliane Segers, and Ludo Verhoeven. 2022.
\newblock The role of phonological awareness, pinyin letter knowledge, and
  visual perception skills in kindergarteners’ chinese character reading.
\newblock \emph{Behavioral Sciences}, 12(8):254.

\bibitem[{Zhou(1958)}]{zhou1958}
Enlai Zhou. 1958.
\newblock 當前文字改革的任務 [current tasks for writing system
  reform].
\newblock \emph{Retrieved December 2, 2022 from
  \href{https://www.marxists.org/chinese/zhouenlai/129.htm}{https://www.marxists.org/chinese/zhouenlai\\/129.htm}}.

\bibitem[{Zhou and Marslen-Wilson(1999)}]{zhou1999sublexical}
Xiaolin Zhou and William Marslen-Wilson. 1999.
\newblock Sublexical processing in reading chinese.
\newblock In \emph{Reading chinese script}, pages 49--76. Psychology Press.

\bibitem[{Ziegler et~al.(2000)Ziegler, Tan, Perry, and
  Montant}]{ziegler2000phonology}
Johannes~C Ziegler, Li~Hai Tan, Conrad Perry, and Marie Montant. 2000.
\newblock Phonology matters: The phonological frequency effect in written
  chinese.
\newblock \emph{Psychological Science}, 11(3):234--238.

\end{thebibliography}
\bibliographystyle{acl_natbib}

\section*{Appendix A. Chinese character naming experiment}
The human Chinese character naming experiment has received IRB approval. The participants were recruited online and in person. The selection criteria include: 1) native speaker of Mandarin; 2) able to read and write in traditional Chinese scripts and pinyin. The participants completed an online questionnaire on qualtrics on their phones or computers.  

The participants were first asked to complete the screening questions to make sure that they are able to read and write in traditional Chinese scripts and pinyin. The screen questions are:\\
\begin{enumerate}
\item[$\bullet$] 請將下面一段漢語拼音翻譯成漢字（聲調用數字代替）(Please transcribe the following pinyin into Chinese. Use numbers to represent the tone)：\\
 `chun1 mian2 bu4 jue2 xiao3, chu4 chu4 wen2 ti2 niao3.' 
    \item[$\bullet$] 請將下面一段漢字翻譯成漢語拼音（聲調用數字代替）(Please transcribe the following Chinese into pinyin. Use numbers to represent the tone)：\\
    `夜來風雨聲，花落知多少。'
\end{enumerate}

Then the participants were asked to provide the pinyin for 60 test characters. The participants first selected `yes' or `no' whether they know the character. Then they were asked to type the pinyin of the character. Example questions are:

\begin{enumerate}
    \item[$\bullet$] 請問您認識 "紑" 這個字嗎？(Do you know the character 紑?) \\
    $\Box$ 認識 (yes) $\Box$ 不認識 (no)
    \item[$\bullet$] 請猜測並標註 "紑" 的讀音：(Please guess and write the pinyin of 紑)
\end{enumerate}

The 60 test characters were separated into 2 test blocks, with 30 characters each. In between the 2 blocks, we set a block of 15 frequent characters and ask the participants to provide the answer to make it more engaging for the participants. An example question is:
\begin{enumerate}
    \item[$\bullet$] 請標註 “河” 的讀音：(Please write the pinyin of “河”)
\end{enumerate}

\section*{Appendix B. Tables of statistic summaries}

\begin{table}[!ht]
\small
\centering
\begin{tabular}{llllll}
\toprule
data & label & -T-S & -T+S & +T-S & +T+S \\
\midrule
\multirow{5}{*}{\dall} & \base & 49.5 & 50.3 & 51.0 & 49.2 \\
 & \labm & 48.2 & 49.0 & 49.8 & 48.8 \\
 & \labs & 48.2 & 50.8 & 50.8 & 51.2 \\
 & \labmr & 52.3 & 53.5 & 53.7 & 50.0 \\
 & \labsr & 51.0 & 51.0 & 52.3 & 49.2 \\
 \midrule

\multirow{5}{*}{\dmid} & \base & 47.0 & 43.7 & 45.3 & 46.7 \\
 & \labm & 47.7 & 48.0 & 48.0 & 45.0 \\
 & \labs & 49.7 & 49.3 & 48.7 & 43.0 \\
 & \labmr & 46.0 & 45.0 & 45.3 & 47.7 \\
 & \labsr & 49.0 & 52.0 & 49.7 & 45.3 \\
 \midrule

\multirow{5}{*}{\dhigh} & \base & 40.0 & 41.3 & 39.7 & 41.3 \\
 & \labm & 39.3 & 39.7 & 42.0 & 39.0 \\
 & \labs & 40.7 & 40.0 & 41.7 & 42.7 \\
 & \labmr & 45.0 & 43.0 & 43.0 & 41.7 \\
 & \labsr & 44.0 & 43.7 & 42.0 & 44.0 \\
 \midrule

\multirow{5}{*}{\begin{tabular}[c]{@{}l@{}}\textsc{all+}\\ \textsc{freq}\end{tabular}} & \base & 48.0 & 49.7 & 52.3 & 49.3 \\
 & \labm & 46.0 & 47.3 & 48.0 & 49.3 \\
 & \labs & 47.3 & 52.7 & 53.3 & 51.0 \\
 & \labmr & 52.3 & 55.0 & 53.7 & 50.3 \\
 & \labsr & 49.7 & 49.3 & 50.7 & 47.3 \\
 \bottomrule
\end{tabular}
\caption{\label{exp2_tab:3_dfreq}The average accuracy (over 5 seeds) on test set for models in Experiment 2 (\MODB) trained on \dhigh, \dmid, or adding frequency label as input features on \dall. {+}T, {-}T, {+}S, {-}S refers to adding tone, no tone, shuffling, and no shuffling, respectively.}
\end{table}
\begin{table*}[!ht]
\centering
\small
\begin{tabular}{lllllll}
\toprule
Label & Freq. & \# of characters & \regular (\%) & \alliterating (\%)& \rhyming (\%)& \rad (\%) \\
\midrule
Rare & = 1 & 1025 & 41.6 & 7.0 & 22.7 & 28.7 \\
Low & 2 - 29 & 1116 & 42.7 & 8.0 & 24.8 & 24.5 \\
Mid & 30 - 2337 & 1070 & 44.6 & 7.5 & 24.1 & 23.8 \\
High & $>$ 2337 & 1070 & 42.1 & 8.7 & 22.5 & 26.7 \\
\bottomrule
\end{tabular}
\caption{The summary of characters in \dallf. } 
\label{app_tab:3_freq}
\end{table*}

\begin{table*}[!ht]
\small
\centering
\begin{tabular}{l@{\hspace{1.2\tabcolsep}}l@{\hspace{1\tabcolsep}}l@{\hspace{1.1\tabcolsep}}ll@{\hspace{1.1\tabcolsep}}l|l@{\hspace{1.1\tabcolsep}}ll@{\hspace{1.1\tabcolsep}}l}
\toprule
 \multicolumn{2}{l}{Overlap Rate}& \multicolumn{4}{c}{No Tone} & \multicolumn{4}{c}{Tone} \\
 &  & \multicolumn{2}{c}{No Shuffle} & \multicolumn{2}{c}{Shuffle} & \multicolumn{2}{c}{No Shuffle} & \multicolumn{2}{c}{Shuffle} \\\midrule
Data & Model & \makecell{\textsc{model}\\ \scriptsize\textsc{[-pinyin]}} & \makecell{\textsc{model}\\ \scriptsize\textsc{[+pinyin]}}  & \makecell{\textsc{model}\\ \scriptsize\textsc{[-pinyin]}} & \makecell{\textsc{model}\\ \scriptsize\textsc{[+pinyin]}}& \makecell{\textsc{model}\\ \scriptsize\textsc{[-pinyin]}} & \makecell{\textsc{model}\\ \scriptsize\textsc{[+pinyin]}} & \makecell{\textsc{model}\\ \scriptsize\textsc{[-pinyin]}} & \makecell{\textsc{model}\\ \scriptsize\textsc{[+pinyin]}}\\
 \midrule
 \dall & \base & 0.43\std{0.06} & 0.47\std{0.06}  &0.44\std{0.06} & 0.45\std{0.06} & 0.41\std{0.06} & 0.47\std{0.06} &0.42\std{0.06} & 0.48\std{0.06} \\
 & \labm &0.41\std{0.06} &0.49\std{0.06} & 0.43\std{0.05} & 0.49\std{0.07} &0.42\std{0.05} & 0.49\std{0.07} & 0.42\std{0.06} & 0.48\std{0.06}\\
 & \labmr & 0.43\std{0.07} & 0.45\std{0.05} & 0.43\std{0.05}& 0.47\std{0.06} & 0.46\std{0.06} & 0.43\std{0.06} & 0.44\std{0.06} & 0.45\std{0.07}  \\
 & \labs & 0.42\std{0.06} & 0.47\std{0.06} & 0.44\std{0.06}& 0.45\std{0.06} & 0.44\std{0.06} & 0.48\std{0.06} & 0.45\std{0.06} & 0.48\std{0.06} \\
 & \labsr & 0.42\std{0.07} & 0.46\std{0.07} & 0.42\std{0.06} & 0.47\std{0.06} &  0.44\std{0.06} & 0.48\std{0.06} &  0.44\std{0.05} & 0.47\std{0.06} \\
 \midrule
 \dmid & \base & 0.41\std{0.06} & 0.46\std{0.07} & 0.41\std{0.06} & 0.44\std{0.06} & 0.39\std{0.06} & 0.45\std{0.07} & 0.41\std{0.06} & 0.45\std{0.07} \\
 & \labm & 0.42\std{0.06} & 0.46\std{0.06} & 0.39\std{0.06} & 0.47\std{0.06} & 0.42\std{0.07} & 0.47\std{0.06} & 0.41\std{0.06} & 0.44\std{0.07} \\
  & \labmr & 0.42\std{0.06} &0.44\std{0.08} & 0.41\std{0.06}& 0.47\std{0.06} & 0.41\std{0.06} & 0.45\std{0.06} & 0.42\std{0.06} & 0.46\std{0.06} \\
 & \labs & 0.41\std{0.05} & 0.48\std{0.06} & 0.42\std{0.06} & 0.47\std{0.07} & 0.37\std{0.06} & 0.48\std{0.07} & 0.40\std{0.07} & 0.46\std{0.07} \\
 & \labsr &0.38\std{0.07} & 0.45\std{0.08} & 0.42\std{0.06}& 0.47\std{0.06} & 0.41\std{0.06} & 0.45\std{0.06} & 0.39\std{0.06} & 0.45\std{0.06}  \\
 \midrule
\dhigh & \base & 0.32\std{0.06} & 0.38\std{0.07} & 0.31\std{0.06} & 0.39\std{0.06} & 0.30\std{0.05} & 0.41\std{0.08} & 0.32\std{0.06} & 0.42\std{0.06}  \\
 & \labm & 0.32\std{0.06} & 0.41\std{0.08} & 0.30\std{0.06} & 0.40\std{0.06} & 0.31\std{0.06} & 0.42\std{0.06} & 0.32\std{0.06} & 0.39\std{0.07}\\
  & \labmr & 0.34\std{0.07}& 0.45\std{0.07} & 0.32\std{0.06} & 0.43\std{0.08} & 0.33\std{0.07} & 0.45\std{0.07} & 0.32\std{0.06} & 0.44\std{0.07} \\
 & \labs & 0.30\std{0.07} & 0.41\std{0.07} & 0.33\std{0.06} & 0.41\std{0.07} &0.31\std{0.06} & 0.43\std{0.06} & 0.34\std{0.06}& 0.43\std{0.07} \\
 & \labsr & 0.31\std{0.05} & 0.45\std{0.06} &0.32\std{0.06}& 0.43\std{0.06} & 0.31\std{0.05} & 0.44\std{0.06} & 0.32\std{0.06}& 0.44\std{0.07} \\
 \midrule
 
\textsc{all+} & \base &  0.43\std{0.06} & 0.46\std{0.07} & 0.42\std{0.06}& 0.45\std{0.07} & 0.44\std{0.06} & 0.47\std{0.06} & 0.43\std{0.06} & 0.47\std{0.06}\\
\textsc{freq} & \labm & 0.44\std{0.06} &0.44\std{0.07} & 0.44\std{0.06} & 0.46\std{0.06} & 0.43\std{0.06} & 0.47\std{0.06} & 0.40\std{0.06}& 0.47\std{0.07} \\
  & \labmr & 0.44\std{0.06} & 0.45\std{0.07} & 0.42\std{0.05}& 0.44\std{0.06} & 0.43\std{0.06} & 0.43\std{0.05} &0.43\std{0.06} & 0.47\std{0.06} \\
 & \labs & 0.43\std{0.06} & 0.46\std{0.06} & 0.44\std{0.06} & 0.48\std{0.06} & 0.42\std{0.06} & 0.50\std{0.06} & 0.41\std{0.06}& 0.48\std{0.06} \\
 & \labsr & 0.43\std{0.05}& 0.46\std{0.06} & 0.43\std{0.06} & 0.47\std{0.06} & 0.41\std{0.07} & 0.46\std{0.06} & 0.45\std{0.06} & 0.45\std{0.06} \\
 \bottomrule
\end{tabular}
\caption{The overlap rate averaged over 275 pairs of answers (5 random seeds x 55 participants) for each model with different labels and conditions.}
\label{app_tab: overlap}
\end{table*}

\end{CJK*}

\clearpage
\newpage

\begin{figure*}
    \centering
    \includegraphics[scale = 0.35]{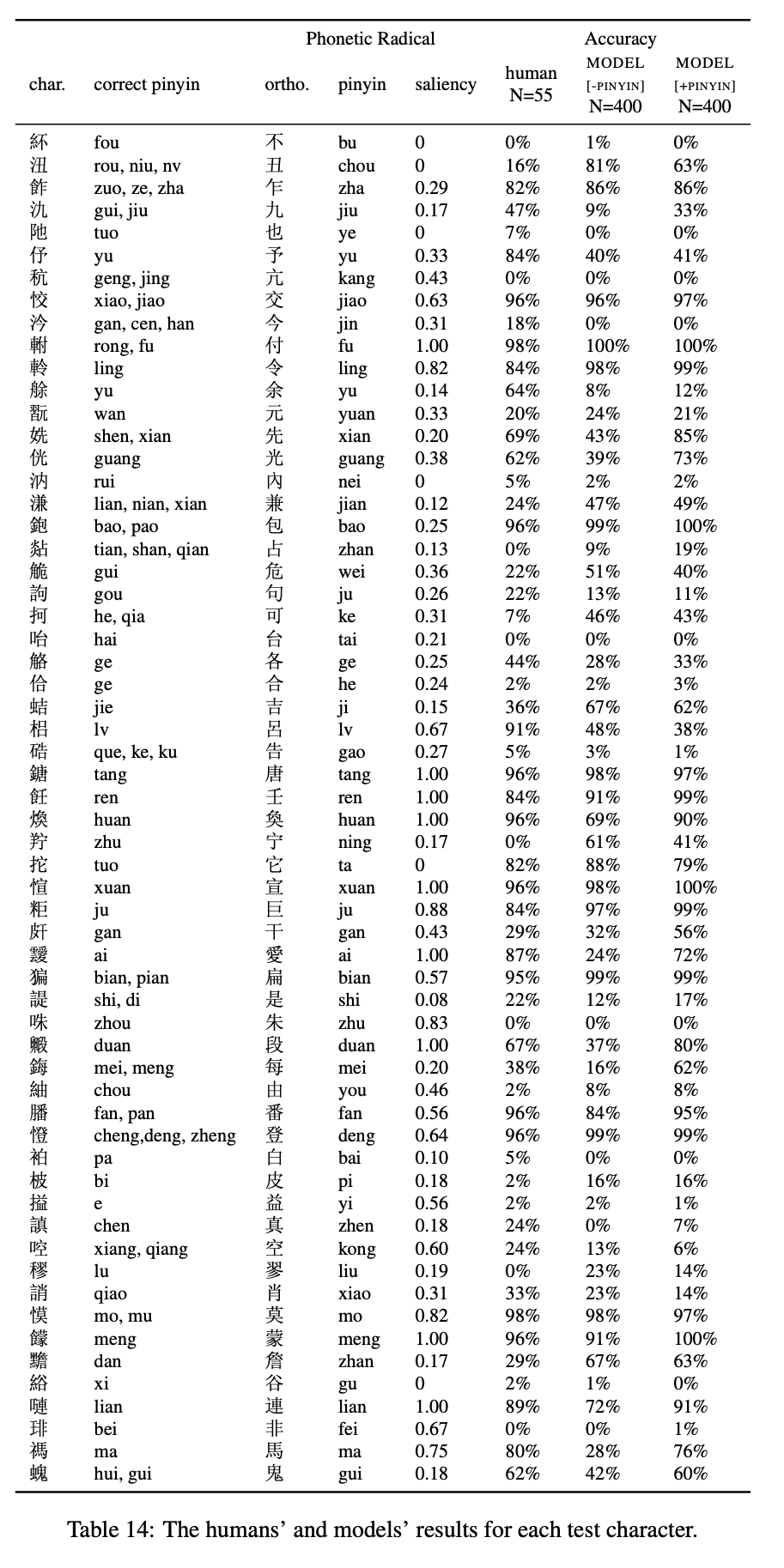}
\end{figure*}
\end{document}